\documentclass[letterpaper]{article} 
\usepackage{aaai25}  
\usepackage{times}  
\usepackage{helvet}  
\usepackage{courier}  
\usepackage[hyphens]{url}  
\usepackage{graphicx} 
\usepackage{natbib}  
\usepackage{caption} 
\usepackage{booktabs}
\usepackage{amsmath}
\usepackage{tabularx}
\usepackage{multicol}
\usepackage{multirow}
\usepackage{amssymb}
\usepackage{algorithm}
\usepackage{algpseudocode}
\usepackage{utfsym}
\usepackage{newfloat}
\usepackage{listings}
\DeclareCaptionStyle{ruled}{labelfont=normalfont,labelsep=colon,strut=off} 
\floatstyle{ruled}
\newfloat{listing}{tb}{lst}{}
\floatname{listing}{Listing}
\title{Enhancing Multimodal Large Language Models Complex Reason via Similarity Computation}
\author {
    Xiaofeng Zhang \equalcontrib \thanks{Corresponding author}\textsuperscript{,\rm 1},
    Fanshuo Zeng \equalcontrib \textsuperscript{,\rm 2},
    Yihao Quan \textsuperscript{\rm 3},
    Zheng Hui \textsuperscript{\rm 4},
    Jiawei Yao \textsuperscript{\rm 5}
}
\affiliations {
    \textsuperscript{\rm 1}Shanghai Jiaotong University\\
    \textsuperscript{\rm 2} Institute of Automation, Chinese Academy of Sciences \\
    \textsuperscript{\rm 3} Beijing Jiaotong University\\
    \textsuperscript{\rm 4} Columbia University\\
    \textsuperscript{\rm 5} University of Washington \\
     {{framebreak@sjtu.edu.cn}}\quad

}

\begin{document}

\maketitle

\begin{abstract}
Multimodal large language models have experienced rapid growth, and numerous different models have emerged. The interpretability of LVLMs remains an under-explored area. Especially when faced with more complex tasks such as chain-of-thought reasoning, its internal mechanisms still resemble a black box that is difficult to decipher. By studying the interaction and information flow between images and text, we noticed that in models such as LLaVA1.5,  image tokens that are semantically related to text are more likely to have information flow convergence in the LLM decoding layer, and these image tokens receive higher attention scores. However, those image tokens that are less relevant to the text do not have information flow convergence, and they only get very small attention scores. To efficiently utilize the image information, we propose a new image token reduction method, Simignore, which aims to improve the complex reasoning ability of LVLMs by computing the similarity between image and text embeddings and ignoring image tokens that are irrelevant and unimportant to the text. Through extensive experiments, we demonstrate the effectiveness of our method for complex reasoning tasks. 
\begin{links}
\link{Code}{https://github.com/FanshuoZeng/Simignore}
\end{links}

\end{abstract}

\section{Introduction}

Large Vision Language Models (LVLMs) have rapidly developed in recent years and become a research hotspot in the field of computer vision as well as natural language processing. Multimodal large language models (LLMs) have shown impressive performance on complex reasoning by leveraging chain-of-thought (CoT) prompting to generate intermediate reasoning chains as the rationale to infer the answer. LVLMs utilizes a visual encoder such as CLIP to process picture patches to get visual tokens, which are used as the context of visual information to accomplish visual-textual reasoning tasks. The visual coder processes these image patches into hundreds of tokens, e.g., 576 for CLIP \cite{clip} and 729 for siglip \cite{sigmoid}. These extra image tokens lead to an increase in computation, and some studies \cite{chu2023mobilevlm, yuan2023tinygpt} have reduced the inference cost by using smaller LLMs with fewer parameters, however, these approaches also lead to a decrease in the inference power of LLMs \cite{chu2024mobilevlm}. Therefore, a better approach is to reduce the computational cost by decreasing the length of input tokens. At the same time, studies \cite{llava-prumerge,fastv} have shown that not all image tokens are important for model inference and that a proper reduction of image tokens can improve the inference ability of LLMs, leading to a significant increase in the accuracy of LLMs on visual complex reasoning tasks.

\begin{figure}[t]
\centerline{\includegraphics[scale=0.20]{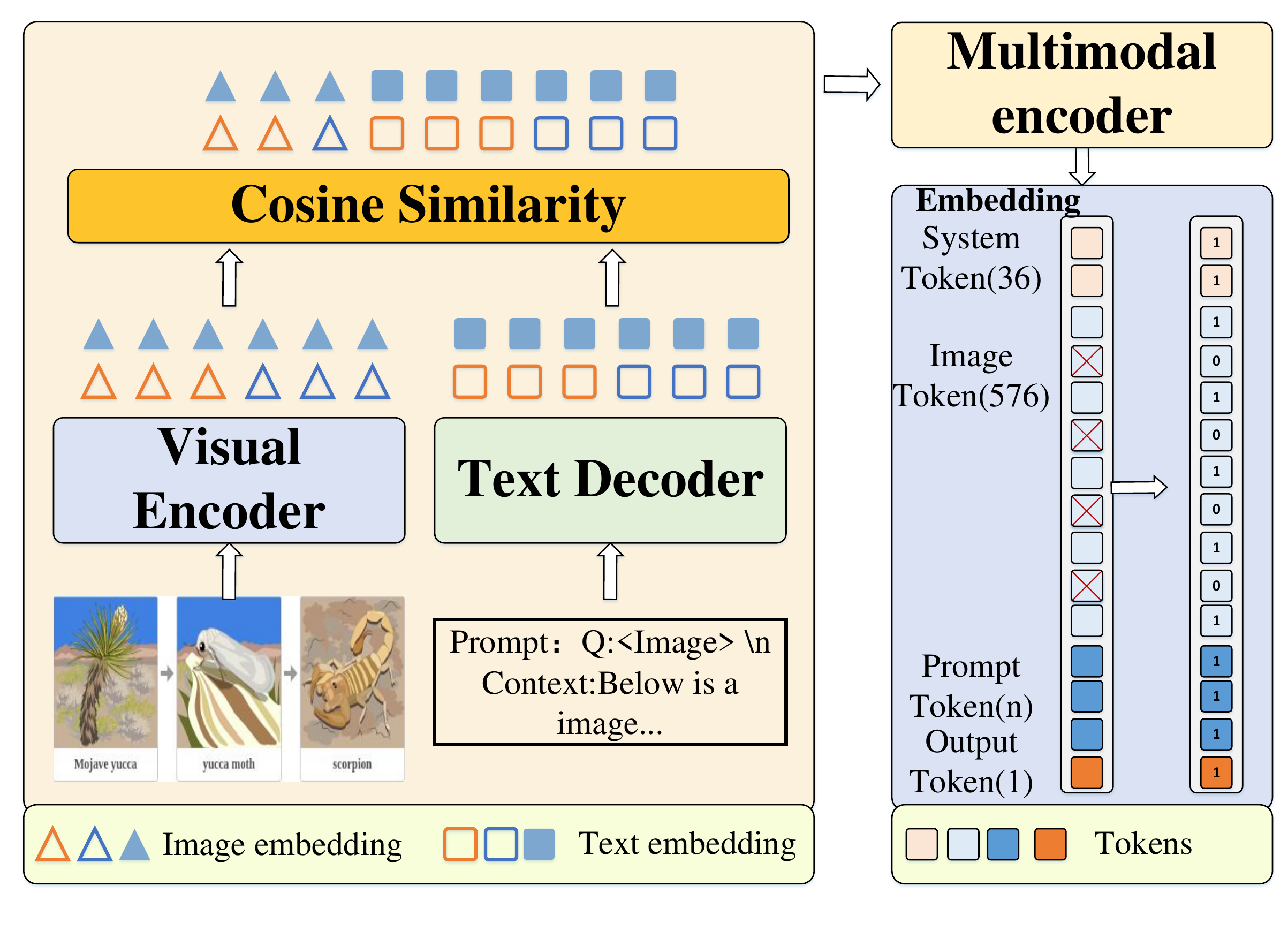}}
\caption{The simplified structure of our method.}
\label{structure}
\end{figure}

There are some works on reducing image tokens, FastV \cite{fastv} visualized the attention scores of system tokens, image tokens, and user tokens during the LLM inference process and found that the weight of image tokens was very small after the second layer, so he chose to discard half of the image tokens with lower scores at the second layer based on the ordering of their attention scores. LLaVA-PruMerge \cite{llava-prumerge} utilized the self-attention mechanism between category tokens and visual tokens to observe the distribution of attention between them and found that most of the visual tokens had attention values close to zero with the category tokens, indicating that these tokens are not critical in the image representation. Therefore he selects appropriate image tokens based on the spatial similarity between visual tokens and CLS tokens, then clusters the pruned tokens and maintains the completeness and richness of the visual information by merging the clustered tokens with the unpruned ones. 

Regarding the above two methods of reducing image tokens, although the two methods of image markup approximation have achieved good results, there are limitations to their methods: (1) the above two methods of reducing image tokens do not take into account the interaction with textual prompt, and it is quite easy to filter out the tokens related to the “semantics” of the text. (2) At present, the interaction between image tokens and text tokens is still unclear, and how LLM utilizes image tokens for answering is still unknown and needs further exploration. 

\begin{figure}[h]
\centerline{\includegraphics[scale=0.22]{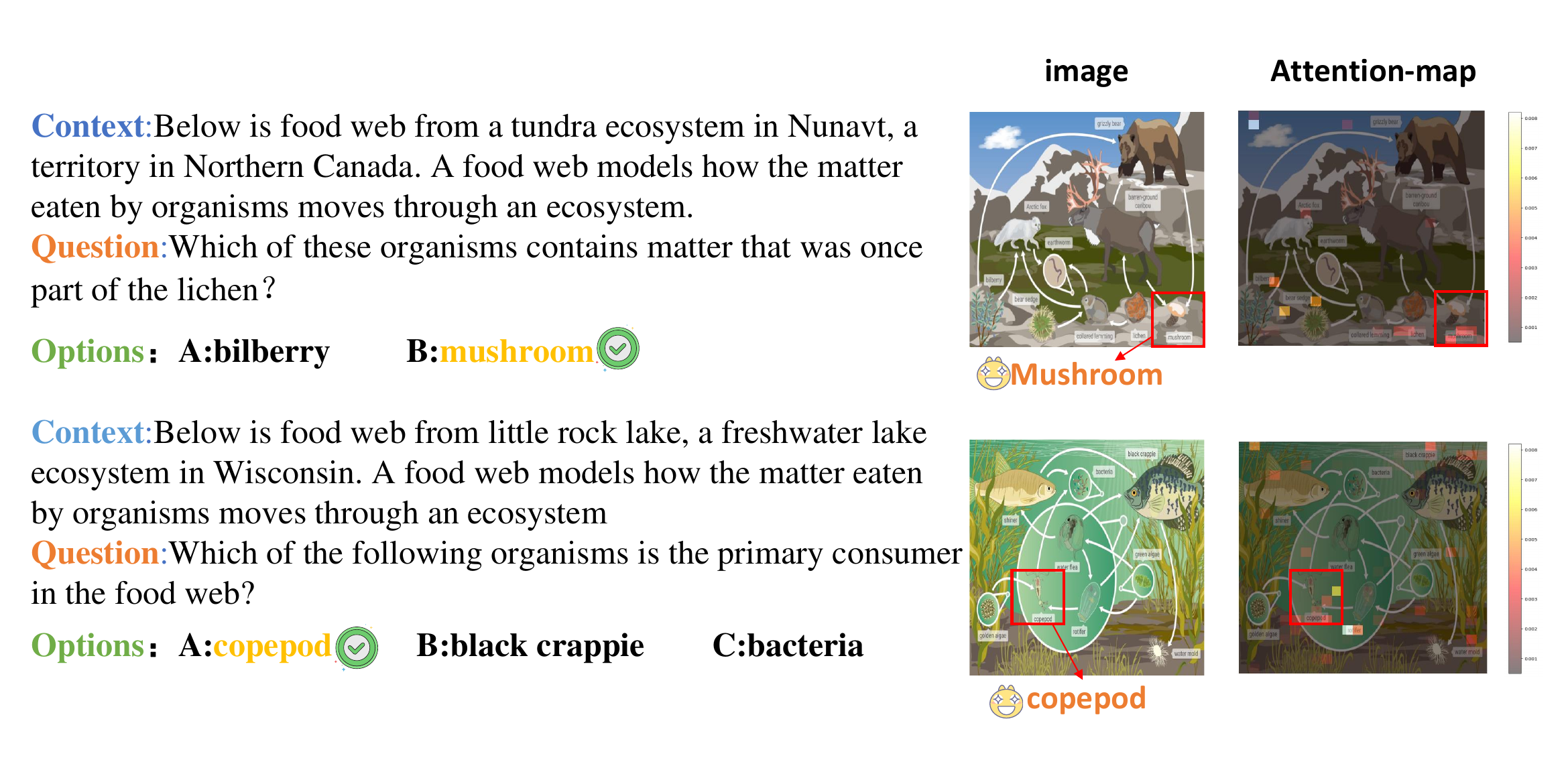}}
\caption{We find that the information flow converges in regions related to the option of prompt, such as mushroom and copepod.}
\label{introduction}
\end{figure}

We employ information flow to explore the interaction between image tokens and text prompts. We define information flow as the process by which image tokens gradually converge on the semantics associated with text during processing in the attention mechanism of image tokens. Specifically, we visualize the attention score of the image token in the LLM decoder and superimpose it on the image patch. We find a convergence of information flow on the image patch related to the text. As shown in Fig.~\ref{introduction}, we can observe that the image patches related to the text options, such as mushroom and bilberry, converge information flow in the network, and they get higher attention scores. In other cases, when answering a question does not require a reference image, the network does not pay much attention to the image and the information flow does not converge. 

Therefore, we design the image-text token filtering algorithm named Simignore to keep irrelevant tokens from causing interference. Specifically, as shown in Fig. \ref{structure}, we map the embeddings of image tokens and prompt tokens to the same similarity metric space, in which all tokens are points on a two-dimensional plane. Then, we compute the similarity value between the image token and text token using the similarity algorithm, we select the K image tokens with the highest similarity and record their subscripts. Finally, we keep the selected image tokens, and for the unselected tokens, we set their corresponding attention mask to 0 to ignore them.

In summary, the contribution of this work is as follows: 
\begin{itemize}
\item We found through information flow that in the LLM decoder, images that are semantically related to textual markups are more likely to have a convergence of information flow.
\item We propose a new method, Simignore, which aims to retain the image tokens that interact with the text and ignore the irrelevant and unimportant image tokens to improve the ability of LVLMs in complex reasoning tasks. 
\item Through extensive experiments, we demonstrate the effectiveness of our method for complex reasoning in visual reasoning tasks with different LVLMs. 
\end{itemize}

\section{Related Work}
\subsection{Multimodal Large Language Models}

Multimodal large language models can process and understand data from many different modalities, such as text, and images, and thus demonstrate excellent performance in a variety of tasks.CLIP~\cite{clip} and BLIP~\cite{blip} use a pre-training approach that maps images and text into the same feature space, enabling the model to correlate images and text by comparing feature vectors. LLaVA~\cite{llava1.5}, MiniGPT-4~\cite{minigpt-4}, Qwen-VL~\cite{qwen-vl}, CogVLM~\cite{cogvlm} and other methods \cite{sun1,sun2,sun3,sun4,tang1,tang2,tang3,tang4,tang5,m1,m2,m3,m4} use pre-trained ViT to process information from images. EAH \cite{zhang1,zhang2,zhang3,zhang4} firstly explained the VLM in a black box through the Angle of information flow, and then found that the information flow was negatively correlated with the hallucination, so the attentional head enhancement method was proposed to alleviate the hallucination. Recent studies have demonstrated the potential of multi-modal learning in applications such as document restoration \cite{xing1,wang1,wang2}, medical image translation \cite{xing2}, missing modality prediction \cite{xing3}, emotion recognition \cite{xing4} and multimodal generation \cite{shen1,shen2,shen3,shen4}.



\subsection{Image-Text Multimodal Similarity}

Image-text similarity research is a key area in multimodal learning for assessing the consistency between images and text, and the accuracy of cross-modal retrieval has been significantly improved in recent years by a variety of methods. These methods include the joint embedding technique~\cite{DeViSE, RKiros, vse++}, the SGRAF network proposed by Diao~\cite{Diao-Similarity}. the instance comparison embedding method by Zeng~\cite{zeng2024image}. the unified comparison learning method by Yang~\cite{yang2022unified}. and the image feature and class semantic embedding in cosine metric space by Liu~\cite{liu2021goal}. In particular, cosine similarity is used to recognize image tokens with strong relevance to textual information due to its advantages in reducing intra-class variance and improving recognition ability.

\begin{figure*}[h]
\centerline{\includegraphics[scale=0.19]{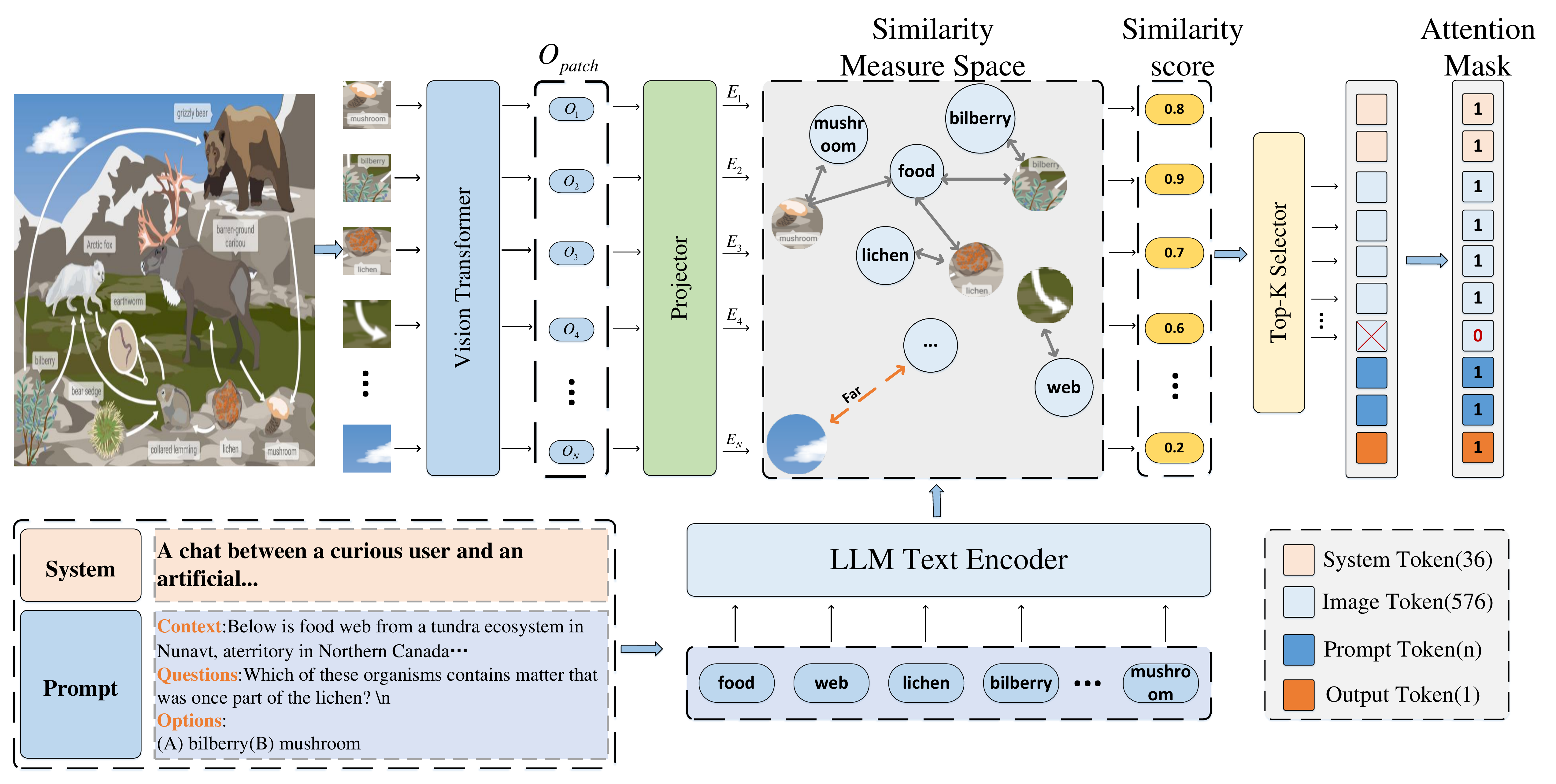}}
\caption{The holistic framework for an approach to enhance complex reasoning in multimodal large language models through similarity computation between image and text embeddings. We map the embeddings of image token and prompt token to the same similarity metric space, a process that involves operations such as regularization. Here we compute their similarity values. Then we select the $K$ image tokens with the highest similarity and consider them important. For unselected tokens, we ignore them by setting their attention mask to $0$.}
\label{holistic-structure}
\end{figure*}

\section{Method}

To present a visualization of the information flow, we employ the Attention Score techniques to gain a comprehensive understanding of the information flow. The Attention Score reveals the forward processing of the model, showing the contribution of different input elements to the final output.

\subsection{Influence Rate of Image Token on Output Token}

For the output token of the complex reasoning task such as the ScienceQA dataset \cite{sqa}, in the $n$-th layer, we define $\mathcal{G}$ as the indices set of all tokens and $\mathcal{G}$ can be divided into three parts that represent the indices set of system, image, and user tokens:
\begin{equation}
\label{jihe}
    \mathcal{G} = \mathcal{S} + \mathcal{I} + \mathcal{U},
\end{equation}
where $\mathcal{S} = \{ 1, \dots, N_{\mathrm{sys}} \}$ represents the index of system token, $N_{\mathrm{sys}}$ represents the length of system token, $\mathcal{I} = \{ N_{\mathrm{sys}} + 1, \dots, N_{\mathrm{sys}} + N_{\mathrm{img}} \}$ represents the index of image token, $N_{\mathrm{img}}$ represents the length of image token, and $\mathcal{U} = \{ N_{\mathrm{sys}} + N_{\mathrm{img}} + 1, \dots, N_{\mathrm{sys}} + N_{\mathrm{img}} + N_{\mathrm{user}} \}$ represents the index of user token, $N_{\mathrm{user}}$ represents the length of user token. $A_{i,j}$ is defined as the total attention score of the output token's attention on different types of tokens. For the $i$-th query token, the attentions from system, image, and user tokens are summed as $1$:

 \begin{equation}
 \label{eq1}
\sum_{j \in {\mathcal{S}}}A_{i, j}+\sum_{j \in {\mathcal{I}}}A_{i, j}+\sum_{j \in {\mathcal{U}}}A_{i, j}=1,
\end{equation}

\begin{figure}[h]
\centerline{\includegraphics[scale=0.20]{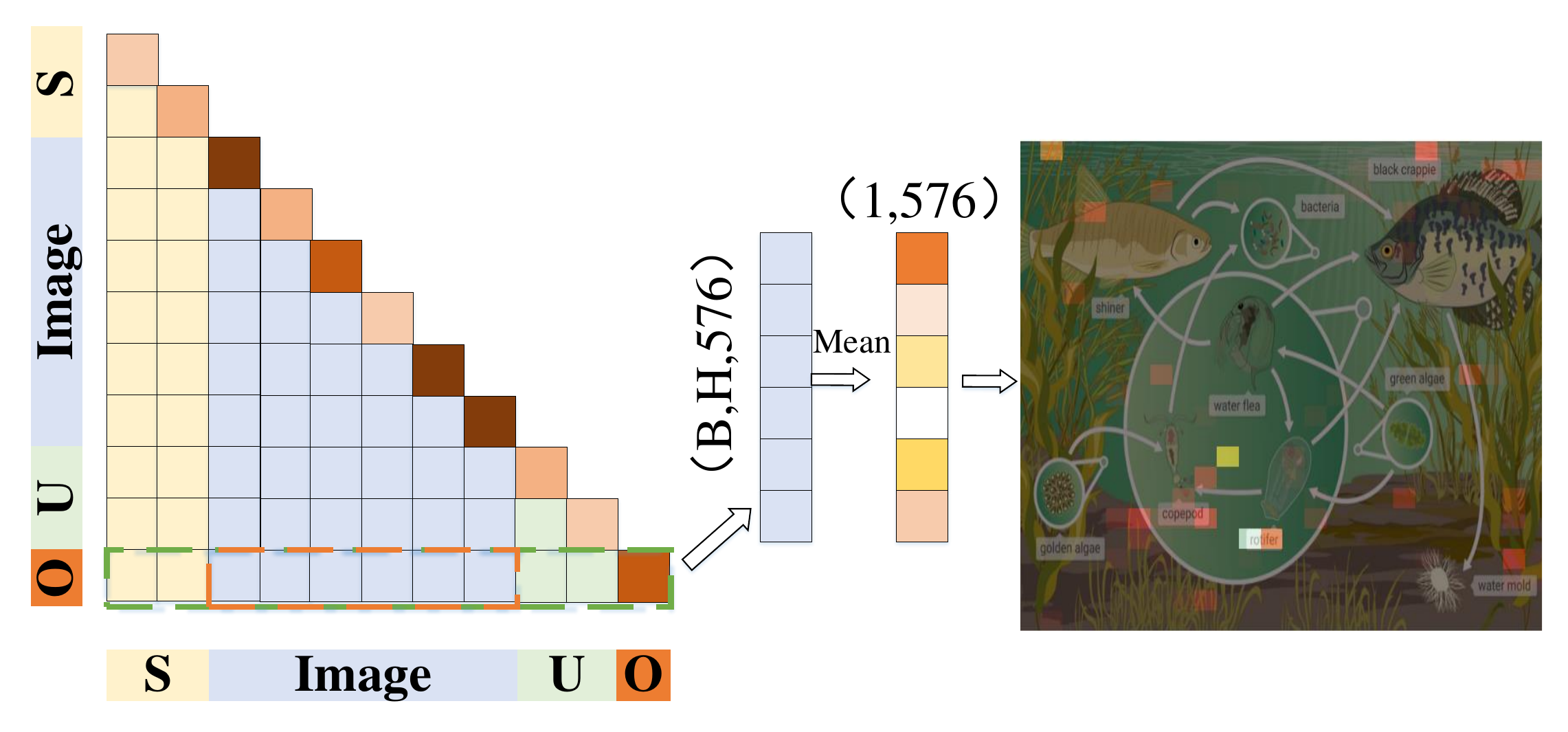}}
\caption{Influence rate of attention score about image tokens.}
\label{Influence rate}
\end{figure}

To ensure that the sum of attention scores for each token is 1, it is necessary to normalize the above summation results to calculate the total attention score for the image token:
\begin{equation}
\lambda_{\mathrm{img}}^j=\sum_{j \in {\mathcal{I}}}A_{i, j} .
\end{equation}
There are 576 image tokens in LLaVA1.5. The shape of the attention mask matrix is $(B, H, N_{img}, N_{img})$, we first perform unsequenced to change the attention matrix to $(H, N_{img}, N_{img})$, the attention matrix is shown in the Fig.~\ref{introduction} and Fig.~\ref{Influence rate}, the horizontal coordinate stands for Q, the vertical coordinate stands for K, and the first row of the vertical coordinate stands for System token, image token, and the influence rate of user token on output token, taking the image token i.e. id from 35-611, we can get the value of 1$\times$576 dimensions, which is the attention score corresponding to 576 tokens, and then we can change the attentions- score reshaped into a 14$\times$14 patch superimposed on the original graph, i.e., the heat map of the influence rate.

Through the influence-score heat map, we can find that the image will have obvious convergence on the options appearing in the prompt, so a way to remove redundancy and keep effective image information is to do the similarity between the image and text token.

\subsection{Compute Similarity between Image and Text Embeddings}
In this section, we will detail our approach to focus on image information related to text. First of all, it is necessary to introduce image embedding and text embedding, and the specific flow is shown in the left part of Fig.~\ref{holistic-structure}.

In the inference process, the input image is \(X \in \mathbb{R}^{C \times H \times W}\), where $C$ denotes the number of channels and $H \times W$ denotes the size of the image. It is then input to the Visual and Language Pre-training model (VLP) for processing:
\begin{equation}
\mathbb{F}: \mathbb{R}^{C \times H \times W} \to \mathbb{R}^{C' \times H' \times W'}
\end{equation}

where $\mathbb{F}$ is the convolution operation, \(C'\) is the new number of channels, \(H'\) and \(W'\) are the new spatial dimensions. Next, the convolved feature mapping is downscaled to a specific size $N_{img} \times I$ using adaptive pooling, function adaptive-pooling $\mathbb{AP}$ is defined:
\begin{equation}
 \mathbb{AP}: \mathbb{R}^{C' \times H' \times W'} \to \mathbb{R}^{N_{img}\times I}
\end{equation}
The whole process is:\\
\begin{equation}
X' = \mathbb{AP}(\mathbb{F}(X))
\end{equation}
where \(X' \in \mathbb{R}^{N_{img} \times I}\) represents the features of the image. \\

The dimension of the text after embedding is \(TextEmb \in \mathbb{R}^{N_{usr} \times T}\). We removed the system tokens from the text and kept only the context, question, and option tokens. $N_{usr}$ denotes the length of the text token and $T$ denotes the dimension of the text embedding. To fuse image information with text information, it is also necessary to embed image and text into the same feature space. So We usually need to align the dimensions of the image features $N_{img} \times I$ with the dimensions of the text features $N_{img} \times T$, function feature-alignment $\mathbb{Fa}$ is defined:
\begin{equation}
\mathbb{Fa}:\mathbb{R}^{N_{img} \times I} \to \mathbb{R}^{N_{img} \times T}
\end{equation}
\begin{equation}
ImgEmb = \mathbb{Fa}(X')
\end{equation}
where \(ImgEmb \in \mathbb{R}^{N_{img} \times T}\) with the same feature dimensions as text. 
At this point, we already have image token and text token embeddings, and next, we describe our approach in detail, as shown in the right part of Fig.~\ref{holistic-structure}. To compute the correlation between image embedding and text embedding, we first normalize $ImgEmb$ and $TextEmb$:

\begin{equation}
ImgEmb_{norm}(i,:) = \frac{ImgEmb(i,:)}{\|ImgEmb(i,:)\|}
\end{equation}
\begin{equation}
TextEmb_{norm}(j,:) = \frac{TextEmb(j,:)}{\|TextEmb(j,:)\|}
\end{equation}
We then compute the cosine similarity matrix:
\begin{equation}
S(i, j)  = ImgEmb_{norm}(i,:)  \\ 
            \cdot TextEmb_{norm}(j,:)^T
\end{equation}
Where \(S \in \mathbb{R}^{N_{img} \times N_{usr}}\), and $S(i, j)$ denotes the similarity between the $i$-th image token and the $j$-th text token. Then, we find the $K$ image tokens with the highest similarity to the text tokens and record their indexes. Specifically, we first need to expand the similarity matrix $S$ into a one-dimensional array              $S_{flat}$ with dimension $N_{img} \times N_{usr}$:

\begin{equation}
S_{flat} = flatten(S)
\end{equation}
Where the dimension of $S_{flat}$ is $1 \times P$, $P = N_{img} \times N_{usr}$. The expanded one-dimensional array $S_{flat}$ is then sorted and the $K$ subscripts with the highest similarity are obtained:
\begin{equation}
Indices = argsort(S_{flat})[-K:]
\end{equation}
Using these one-dimensional subscripts, we can determine the subscripts of the corresponding image token. Since each subscript \( i \) corresponds to a 2D coordinate $(\lfloor i/N_{usr} \rfloor, i \% N_{usr})$, we only need the $i // N_{usr}$ part to represent the subscript of the image token:
\begin{equation}
Indimg_i = \lfloor Indices_i/N_{usr} \rfloor
\end{equation}

Then, we set the attention mask of the tokens that do not belong to $Indimg$ to $0$:


\begin{equation}
\mathbf{M}^{I}_i = \begin{cases}
   1 & \text{if } i \in Indimg \\
   0 & \text{else}
\end{cases}
\end{equation}

Finally, we splice the attention masks of the system token, image token, and user token:
\begin{equation}
\mathbf{M} = \mathbf{M}^{S} + \mathbf{M}^{I} + \mathbf{M}^{U}
\end{equation}
Where $\mathbf{M}^{S}=ones\in \mathbb{R}^{1 \times N_{sys}}$, and $\mathbf{M}^{U}=ones\in \mathbb{R}^{1 \times N_{usr}}$.

\begin{table*}
  \centering
  \resizebox{0.85\textwidth}{!}{%
  \begin{tabular}{c|ccccc}
    \toprule
    Method & Learning & LLM backbone & Res & SQA(IMG)$\%$ \\
    \midrule
    Instruct-BLIP \cite{instructblip} & Zero-shot & Vicuna-7B & 224 & 60.50 \\
    Instruct-BLIP \cite{instructblip} & Zero-shot & Vicuna-13B & 224 & 63.10\\
    BLIP-2 \cite{blip2} & Zero-shot & Vicuna-13B & 224 & 61.03\\
    Shikra \cite{shikra} & Zero-shot & Vicuna-13B & 224 & 45.80\\
    DDCoT(GPT3.5) \cite{ddcot} & Zero-shot & 175B & - & 72.53\\
    DDCoT(MiniGPT-4) \cite{ddcot} & Zero-shot & Vicuna-13B & - & 56.72\\
    Ying-VLM \cite{ying} & Zero-shot & - & - & 55.70\\
    Otter \cite{otter} & Zero-shot & - & - & 66.30\\
    MiniGPT-4 \cite{minigpt-4} & Zero-shot & Vicuna-13B & 336 & 42.34\\
    Qwen-VL-Chat \cite{qwen-vl} & Zero-shot & Qwen-7B & 448 & 68.21\\
    MobileVLM \cite{mobilevlm} & Zero-shot & MobileLLaMA & 336 & 61.00\\
    Qwen-VL \cite{qwen-vl} & Zero-shot & Qwen-7B & 448 & 67.12\\
    \hline
    Mipha-3B \cite{llava1.5} & Zero-shot & Phi-2-2.7B & 384 & 70.40 \\ 
    Mipha-3B+ours & Zero-shot & Phi-2-2.7B & 384 & \textbf{70.85}\\
    LLaVA1.5 \cite{llava1.5} & Zero-shot & Vicuna-7B & 336 & 65.15\\
    LLaVA1.5+ours & Zero-shot & Vicuna-7B & 336 & \textbf{68.02}\\
    LLaVA1.5 \cite{llava1.5} & Zero-shot & Vicuna-13B & 336 & 72.09\\
    LLaVA1.5+ours & Zero-shot & Vicuna-13B & 336 & \textbf{73.23}\\
    \bottomrule
  \end{tabular}}
  \caption{Comparison among different LVLMs on ScienceQA benchmarks, ``Res'' represents the input image resolution.}
  \label{table-compare}
\vspace{-0.3cm}
\end{table*}

\section{Experiment}

\subsection{Dataset and Implementation Detail}


The ScienceQA \cite{sqa} dataset is currently the only dataset available for complex reasoning and contains 21,208 Q\&A multiple-choice questions from elementary and middle school science curricula. A typical question contains multimodal context, correct options, generalized background knowledge, and specific explanations. Our experiments were performed on a single 4090D GPU.

\subsection{Attention Convergence }

When solving visual complex reasoning questions, humans tend to look for information in images that are related to the text. This behavior can efficiently obtain the key information of the image. To explore the degree of attention to image information during multimodal large model reasoning, as shown in Fig.~\ref{introduction}, we obtained the attention scores of image tokens and superimposed them on the original image. Images will have a significant tendency to the options appearing in the cue, specifically, LLM will pay special attention to the word or object in the image that is closely related to the text information. This indicates that image interacts with text in the reasoning process, and LVLMs is more inclined to pay attention to the image tokens that are related to the text.

\begin{figure}[h]
\centerline{\includegraphics[scale=0.3]{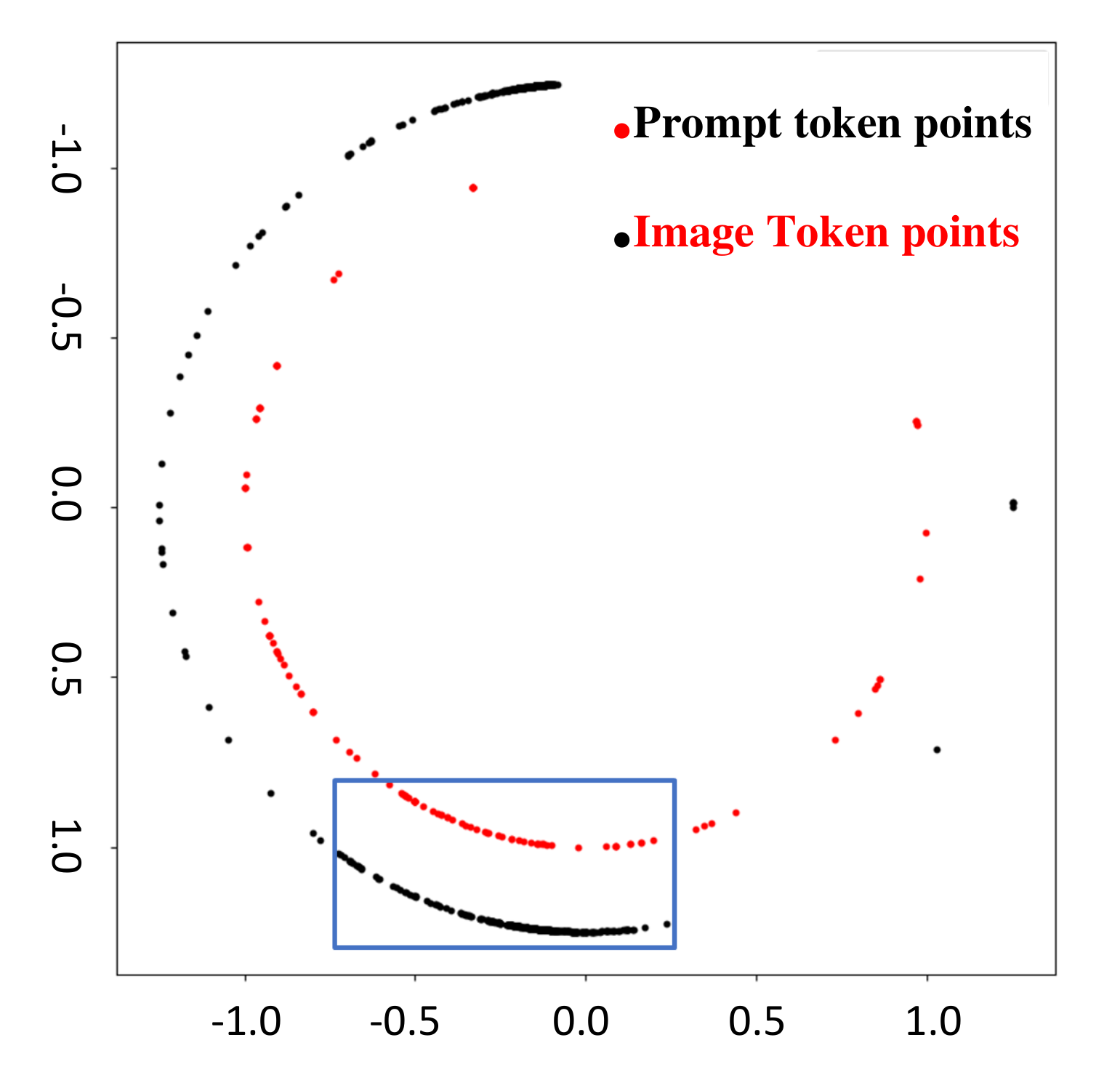}}
\caption{Distribution of image token and prompt token in cosine metric space.}
\label{cosine-metric}
\end{figure}

\subsection{Image-text Dimilarity}
We explored the relationship between image tokens and text tokens. We map embedding to the same cosine metric space, as shown in Fig.~\ref{cosine-metric}, and we find that there is a similarity between the image token and text token in the cosine metric space. Specifically, we find that the distribution of image tokens is similar to that of text tokens, and both of them are concentrated in the metric space of $X < 0$. In the blue rectangle, image tokens and text tokens are densely distributed and have strong similarities. Based on the attention convergence phenomenon and the similarity between image and text in the cosine metric space, we design a method for selecting image tokens with strong relevance to text so that LVLMs obtains useful image tokens and avoids the influence of irrelevant tokens on LVLMs reasoning.

We use different LVLMs for evaluation on the ScienceQA dataset, including the 7b and 13b models of LLaVA-v1.5, and the Mipha-3B model. As shown in Table~\ref{table-compare}, we show the results for the different baseline models as well as for the models to which our method is applied. As can be seen in Table~\ref{table-compare}, the models after applying our method have a great improvement compared to the baseline model, especially the LLaVA1.5-7B model improves the accuracy by 2.87\%. In the experiments with ScienceQA (Image) as the dataset, our method is optimal in both Zero-shot Learning methods with Vicuna-7B or Vicuna-13B as the backbone.

\subsection{Ablation Study}

\subsubsection{Ignore Varying Amounts of Image Tokens}

To explore the effect of ignoring different numbers of image tokens on LLM's complex reasoning task, we set up ten sets of comparison experiments. As shown in Table~\ref{table-number-compare}, we obtained the accuracy as well as the running time of LLM on the ScienceQA (Image) dataset with different numbers of image tokens. Note that the LLM we use is LLaVA1.5-7B. We find that the time for LLM to perform a run is also reduced when some image tokens are ignored. The more image tokens are ignored, the less time is consumed. This indicates that when we ignore some unimportant image tokens, the original dense attention matrix becomes diluted, which greatly reduces the computation. As the number of ignored tokens increases, the attention matrix becomes more sparse and less computationally intensive. Interestingly, when we ignore all the image tokens, the accuracy of LLM still has 53.35\%. However, since it is unlikely that all data have the same image, it is crucial to adaptively choose the appropriate number of image tokens to ignore, which will be our next research work.

\begin{table}
    
    \centering
    \resizebox{0.45\textwidth}{!}{%
    \begin{tabular}{ccc}
    \hline
    \textbf{Ignored number} & \textbf{Accuracy($\%$)} & \textbf{Running time(/s)} \\
    \hline
     72 /576  & 67.23 & 292 \\
     124 /576 & \textbf{68.02} & 279 \\
     144 /576 & 67.67 & 277 \\
     216 /576 & 67.28 & 265 \\
     288 /576 & 66.48 & 260 \\
     360 /576 & 66.04 & 255 \\
     432 /576 & 65.54 & 252 \\
     504 /576 & 63.96 & 246 \\
     576 /576 & 53.35 & 244 \\
     Baseline & 65.15 & 303 \\
    \hline
    \end{tabular}}
    \caption{\label{table-number-compare}
    Accuracy and runtime of LLM when ignoring different numbers of image tokens(baseline: LLaVA1.5-7B).}
\end{table}
In this section, we investigate the effects of ignoring different numbers of image tokens, ignoring image tokens with different levels of importance, and different similarity algorithms on the experimental results. In addition, we also present experiments ignoring text tags.

\subsubsection{The impact of image tokens of different importance on the inference ability}
\label{sec:4-4-2}
To investigate the importance of image tokens in improving the model's complex reasoning ability, we conducted an ablation study. As shown in Table~\ref{table-importance-compare}, we conducted four sets of experiments, namely, ignoring unimportant image tokens (Experiment 1), ignoring image tokens of intermediate importance (Experiment 2), ignoring important image tokens (Experiment 3), and randomly ignoring image tokens (Experiment 4). In particular, the importance of an image token is calculated based on its similarity to text. In Experiment 2, we conducted ten sets of experiments and took the average as the final result. The results show that the best results were obtained in Experiment 1 and poor results were obtained in Experiment 3, indicating that important image tokens play a positive role in the complex reasoning task of LLM, while unimportant image tokens play a negative role in the complex reasoning task of LLM. The results of Experiment 2 are located between Experiment 1 and Experiment 3, indicating that the accuracy of LLM's complex reasoning task shows a positive correlation with the importance of image tokens. The results of Experiment 4 contain both higher and lower scores than Baseline. I believe that lower scores are obtained when some important image tokens are randomly ignored, and higher scores are obtained when some unimportant image tokens are randomly ignored. We provide detailed results of Experiment 4 in supplementary material.

\begin{table}
    \centering
    \resizebox{0.45\textwidth}{!}{%
    \begin{tabular}{cccc}
    \hline
    \textbf{Ignored} & \textbf{Model} & \textbf{Ignored number} & \textbf{Accuracy($\%$)}\\
    \hline
    Unimportant & LLaVA1.5-7B & 124 /576 & \textbf{68.02} \\
    Intermediate & LLaVA1.5-7B & 124 /576 & 64.55 \\
    Important & LLaVA1.5-7B & 124 /576 & 61.73 \\
    Random & LLaVA1.5-7B & 124 /576 & 65.12 \\
    Baseline     & LLaVA1.5-7B & 0 /576   & 65.15 \\
    \hline
    \end{tabular}%
    }
    \caption{\label{table-importance-compare}
    The effect of image tokens of varying importance on a complex reasoning task.}
\end{table}

\begin{figure}[h]
\centerline{\includegraphics[scale=0.19]{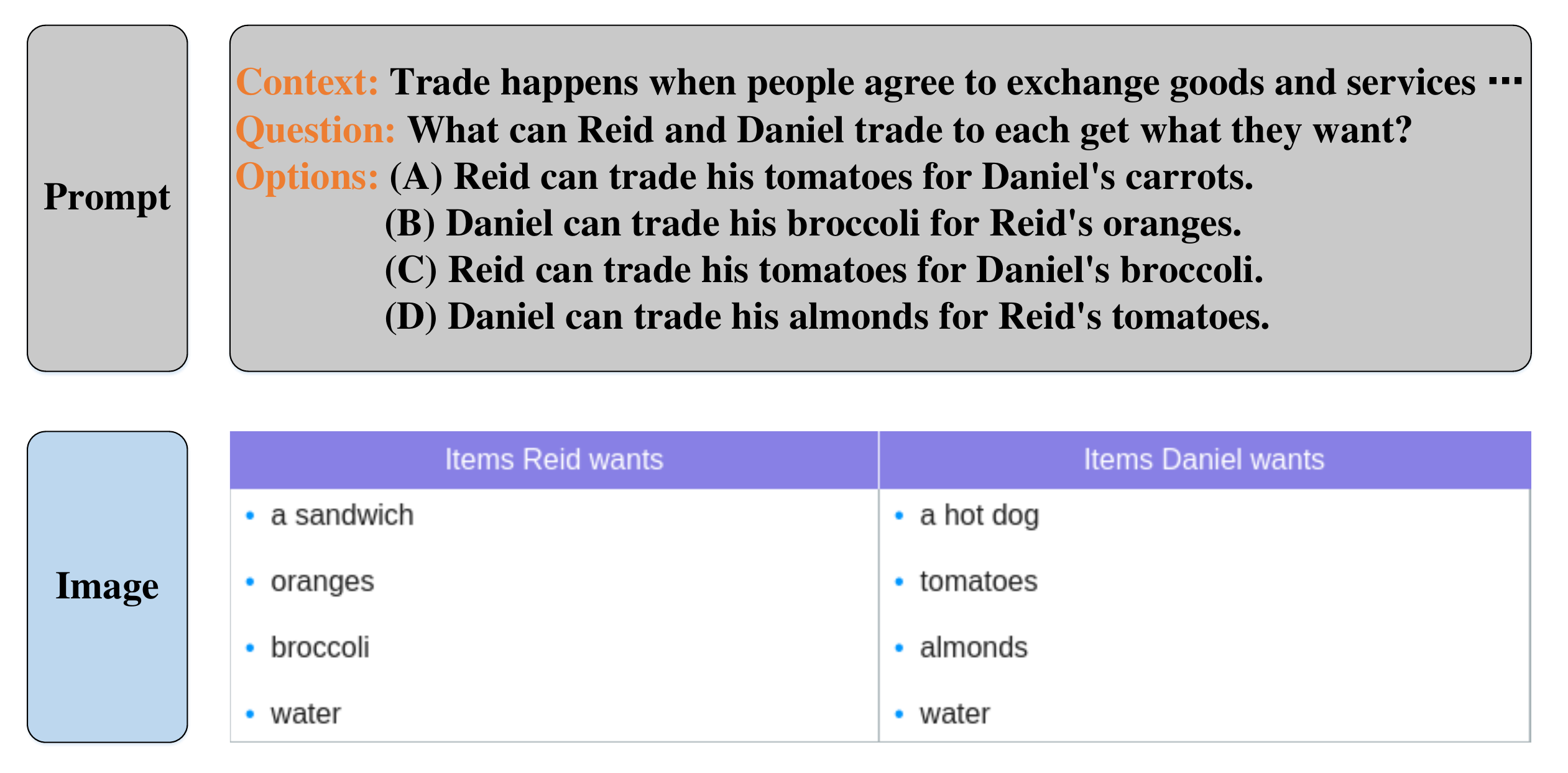}}
\caption{This is a question closely related to the information in the picture and the correct answer to this question is 'C'.}
\label{1879-text-image}
\end{figure}


\begin{figure*}[h]
\centerline{\includegraphics[scale=0.37]{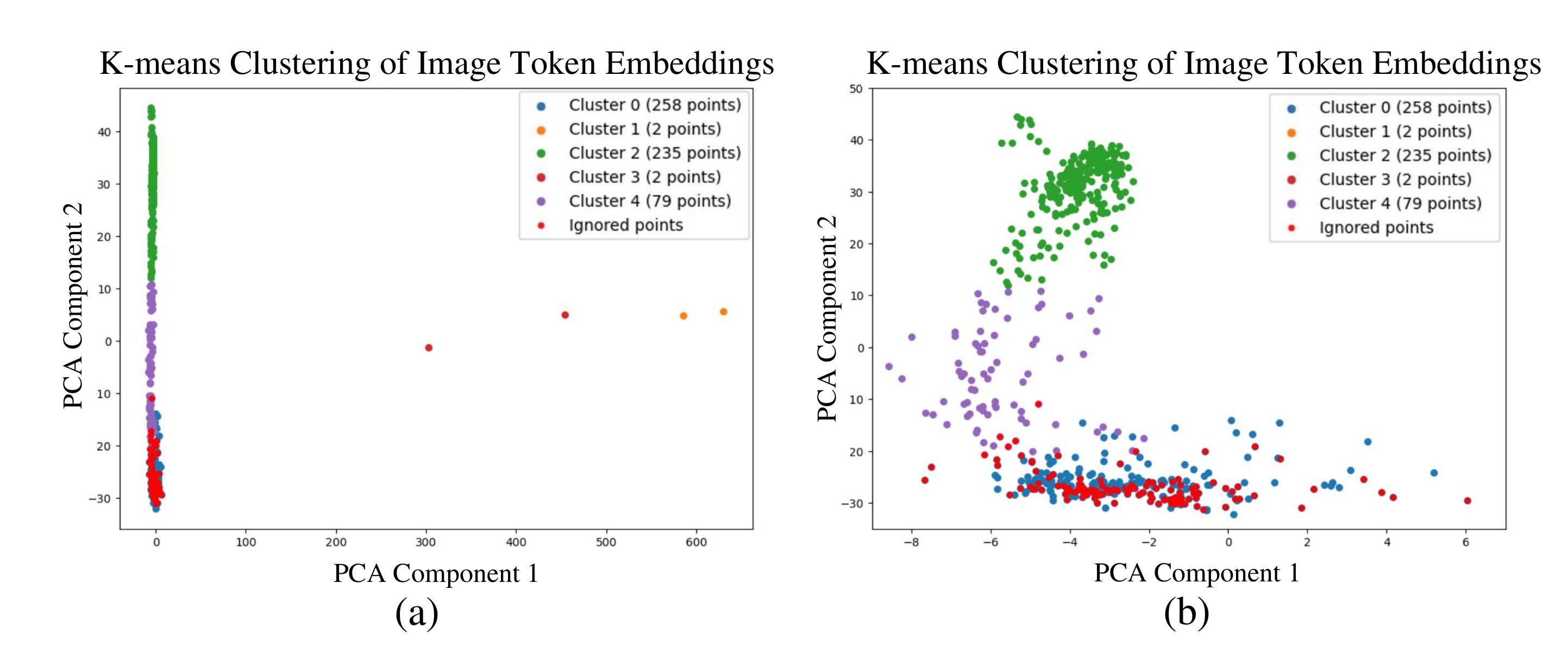}}
\caption{Fig.(a) represents the distribution of image tokens. Where the red dots indicate the distribution position of tokens which are ignored according to the similarity, Fig.(b) is a localized zoomed-in view of Fig.(a).}
\label{1879-all}
\end{figure*}

\subsubsection{Different similarity algorithms}
To investigate the effect of different similarity algorithms on the experimental results, we conducted three different sets of experiments. We conducted experiments using cosine similarity, Euclidean distance similarity, and Manhattan distance similarity, and obtained the accuracy of the LLM when using different schemes. As shown in Table~\ref{table-algorithm-compare}, we found the best improvement using the cosine similarity algorithm, and the other two algorithms also achieved a small improvement. This is because cosine similarity performs better in embedding similarity calculation tasks by measuring the angle between vectors instead of the absolute distance, which ignores scale differences and can better capture the semantic correlation between image and text embedding. In high-dimensional spaces, the distance between vectors tends to be uniform, and Euclidean and Manhattan distances cannot measure this similarity equally effectively because they are affected by the size and scale of the vectors.

\begin{table}[h]
    \centering
    \resizebox{0.45\textwidth}{!}{%
    \begin{tabular}{cccc}
    \hline
    \textbf{Model} & \textbf{Algorithm} & \textbf{Number of ignored} & \textbf{Accuracy} \\
    \hline
     LLaVA1.5-7B & Cosine Similarity  & 124      & \textbf{68.02} \\
     LLaVA1.5-7B & Euclidean Distance & 124      & 66.73 \\
     LLaVA1.5-7B & Manhattan Distance & 124      & 66.88 \\
     Baseline    &  & 0  & 65.15 \\ 
    \hline
    \end{tabular}}
    \caption{\label{table-algorithm-compare}
    The effect of different algorithms for computing similarity on the accuracy of LLM complex reasoning.}
\end{table}

\subsection{Using cluster analysis on image token embedding}

To further explore how our work improves LLM answer accuracy, we performed some analysis on image data. Take the question with id 1879 in the ScienceQA dataset as an example. Fig.~\ref{1879-text-image} shows the prompt of the question with the image. In the baseline method, LLM gave the wrong answer. After using our proposed method, LLM successfully answered the correct answer. To investigate why LLM can correctly infer the answer after ignoring some image tokens, we use k-means clustering for image embedding, where the embedding tensor of each image token is mapped as a point in the 2D plane. Fig.~\ref{1879-all} shows the embedding of the 576 image markers mapped to points on the 2D plane after clustering. The red points indicate the mapped points of the image tokens that we ignore. We notice that the image tokens ignored by our method are concentrated in cluster0, so we ignore all 258 image tokens in cluster0 and still get the correct answer. In addition, we also did a comparison experiment, as shown in Table~\ref{table-clustering-compare}. By analyzing the above experiments, we propose some conjectures: there are some 'spy' tokens in cluster0, which affect LLM's understanding of the image and cause answering errors. Cluster2 and cluster4 contain some important tokens, which will also affect LLM’s understanding of the image if ignored.

To verify our conjecture, we found a critical value: 86. We find that ignoring any number less than 86 causes LVLMs to answer incorrectly while ignoring a number not less than 86 causes LVLMs to answer correctly, and these 86 tokens are all located in cluster0, which suggests that the 86 tokens located in cluster0 will not improve LVLMs's comprehension of the image, but rather play a negative role in the accuracy of LVLMs's answer. We analyzed more data and came to similar conclusions.

\begin{table}[h]
    \centering
    \resizebox{0.45\textwidth}{!}{%
    \begin{tabular}{cccccc}
    \hline
    \textbf{cluster0} & \textbf{cluster1} & \textbf{cluster2} & \textbf{cluster3} & \textbf{cluster4} & \textbf{Result} \\
    \hline
     \usym{2717} & \usym{2717} & \usym{2717} & \usym{2717} & \usym{2717} & F \\
     \usym{2714} & \usym{2714} & \usym{2717} & \usym{2717} & \usym{2717} & T \\
     \usym{2714} & \usym{2717} & \usym{2714} & \usym{2717} & \usym{2717} & T \\
     \usym{2714} & \usym{2717} & \usym{2717} & \usym{2714} & \usym{2717} & T \\
     \usym{2714} & \usym{2717} & \usym{2717} & \usym{2717} & \usym{2714} & T \\
     \usym{2717} & \usym{2714} & \usym{2717} & \usym{2717} & \usym{2717} & F \\
     \usym{2717} & \usym{2717} & \usym{2714} & \usym{2717} & \usym{2717} & F \\
     \usym{2717} & \usym{2717} & \usym{2717} & \usym{2714} & \usym{2717} & F \\
     \usym{2717} & \usym{2717} & \usym{2717} & \usym{2717} & \usym{2714} & F \\
     \usym{2717} & \usym{2717} & \usym{2714} & \usym{2717} & \usym{2714} & T \\
    \hline
    \end{tabular}}
     \caption{\label{table-clustering-compare}
    The answer to the LLM response is obtained by ignoring the image tokens in different clusters. \usym{2714} means ignored, \usym{2717} means do not ignored.}
\end{table}

\section{Discussion and Limitations}

Although Simignore has achieved significant results in improving the inference capabilities of LVLMs, there are still some limitations. Future work will explore strategies for adaptively choosing the number of image tokens to ignore, as well as investigate the internal mechanisms of the model to more fully understand how image and textual information work together to facilitate complex reasoning.

\section{Conclusion}
This study proposes an innovative method to enhance the performance of LVLMs in complex reasoning tasks by computing the similarity between image and text embeddings. We find that in the LLM decoder, image tokens semantically related to text are more likely to converge information flows. Based on this finding, we designed the Simignore algorithm, which improves the similarity between computed images and text embeddings and filters out irrelevant image information, demonstrating that Simignore improves complex reasoning for different LVLMs. 

\bibliography{aaai25}

\end{document}